\documentclass[runningheads]{llncs}

\usepackage{graphicx}
\usepackage{amsmath}
\usepackage{amssymb}
\usepackage{subcaption}
\usepackage[hidelinks]{hyperref}
\usepackage{cite}
\usepackage{tikz}
\usepackage{tikz-3dplot}

\usepackage[absolute,overlay]{textpos}

\begin{document}

\title{Real-time Pose Estimation from Images \\ for Multiple Humanoid Robots}

\author{Arash Amini\inst{1} \and Hafez Farazi\inst{2} \and Sven Behnke\inst{2}}

\authorrunning{Amini, Farazi and Behnke.}

\institute{\textit{University of Bonn, Computer Science Institute VI, Autonomous Intelligent Systems} \\
Friedrich-Hirzebruch-Allee 8, 53115 Bonn, Germany  \\
\email{amini@uni-bonn.de}\inst{1}, \email{\{farazi, behnke\}@ais.uni-bonn.de}\inst{2}}

\begin{textblock*}{20cm}(1cm,1cm) % {block width} (coords) 
\noindent In Proceedings of 24th RoboCup International Symposium, June 2021.\\
Finalist for Best Paper Award
\end{textblock*}

\maketitle              

\begin{abstract}

Pose estimation commonly refers to computer vision methods that recognize people's body postures in images or videos. With recent advancements in deep learning, we now have compelling models to tackle the problem in real-time. Since these models are usually designed for human images, one needs to adapt existing models to work on other creatures, including robots. This paper examines different state-of-the-art pose estimation models and proposes a lightweight model that can work in real-time on humanoid robots in the RoboCup Humanoid League environment. Additionally, we present a novel dataset called the HumanoidRobotPose dataset. The results of this work have the potential to enable many advanced behaviors for soccer-playing robots.

\end{abstract}

\section{Introduction}

\begin{figure}[!ht]
\centering
\includegraphics[width=\linewidth]{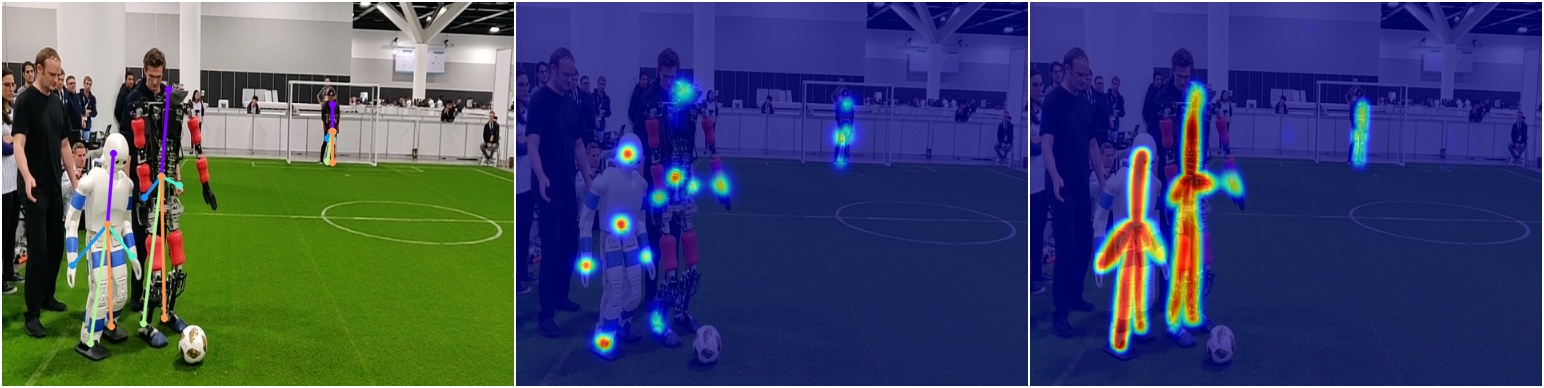}
\caption{One sample image of the introduced dataset with the estimated poses is shown on the left. Our model's predicted heatmaps of all keypoints and limbs are displayed in the middle and right, respectively.}
\label{fig:1}
\end{figure}

The 2D humanoid pose estimation problem aims to detect and localize keypoints and parts and infer the limb connections to reconstruct the existing human poses from images.
The human pose estimation problem's importance arises from the fact that this task has many applications in various areas such as human-computer interaction and action recognition. 
In this work, we address the real-time pose estimation problem for humanoid robots (see Fig.~\ref{fig:1}). The shape similarity between humanoid robots and persons is a double-edged sword. On the one hand, it enables us to start with existing methods designed for persons, but on the other hand, it adds additional difficulty to our problem not to confuse humans with humanoid-robots, especially in adult-size league. In general, the attempts to address the pose estimation problem for multiple persons]can be categorized as either top-down or bottom-up approaches.

In the top-down models, the procedure includes two distinct steps. The first step is to detect individual people, while the particular pose is estimated in the next step. One of these models' disadvantages is that the performance of the model is tightly correlated to the person detector performance. Although the state-of-the-art (SOTA) results are derived from this type of models, including Cascaded Pyramid Network~\cite{Chen2018} and High-Resolution Net (HRNet)~\cite{sun2020}, the runtime of such approaches is negatively affected by the number of persons present, as a single-person pose estimator is run for each detection. Hence, the computational cost linearly increases with more persons, so the performance is often not real-time.

In contrast, bottom-up approaches detect body joints and group them into individuals simultaneously; therefore, they are less dependant on the number of persons in the image. One of the bottom-up method's main challenges is to group the detected keypoints in a real-time manner accurately. Recent approaches~\cite{Cao2021, Papandreou2018, Kreiss2019} utilize a greedy algorithm to group the detected keypoints into individual instances. Moreover, the bottom-up method's performance is more vulnerable to the different scales of the persons in a given image compared to the top-down approaches. To alleviate this issue, previous works exploit the scale search method~\cite{Cao2021} or rely on high-resolution input size~\cite{Papandreou2018}. These solutions are increasing the inference time though. A time-efficient method predicting keypoints at higher resolution was introduced by Cheng et al.~\cite{Cheng2020}, narrowing the performance gap between bottom-up and top-down models.

This paper opts for a bottom-up approach designed for 2D pose estimation of multiple humanoid robots. We made this choice because, in top-down methods, the inference time is generally much higher than in bottom-up approaches, so they will not be suitable for RoboCup real-time applications.
Furthermore, we wanted to avoid the cost of annotating bounding boxes. We remedy our bottom-up model scale variations problem by using feature pyramid structure~\cite{Lin2017} through utilizing high-resolution feature maps.

Despite the availability of several large-scale benchmark datasets such as MPII Human Pose~\cite{Andriluka2014} and MS COCO~\cite{Lin2014COCO} for the task of human pose estimation, we cannot fully utilize them because of differences between robots and humans, such as types and sizes. Thus, we present a new pose dataset of robots from the RoboCup Humanoid League. The code and dataset of this paper are publicly available.\footnote{
\url{https://github.com/AIS-Bonn/HumanoidRobotPoseEstimation}.}
In summary, we make the following contributions:
\begin{itemize}
\item We propose a deep learning model specifically designed to address the 2D pose estimation problem for multiple humanoid robots.
\item We introduce a new dataset, namely the HumanoidRobotPose dataset, consisting of robots from the RoboCup Humanoid League.
\item We demonstrate that the proposed real-time light-weight model outperforms the SOTA bottom-up methods in our application. 
\end{itemize}

\section{Related Work}
Although there are some works on the detection and tracking of humanoid robots~\cite{farazi2017online,farazi2016real}, to the best of our knowledge, there is no previous work that addresses humanoid robot's pose estimation, which works on a variety of robots. Giambattista et al.~\cite{di2019field} propose a gesture-based communication between Nao robot that utilizes OpenPose~\cite{Cao2021} for Nao robot pose estimation. Note that pose estimation on a single standardized Nao robot type is significantly easier than what we need in the Humanoid League. We have to address various unseen robots with different colors, kinematic shapes, and sizes.

\subsubsection{Top-down:} 
Most of the existing top-down methods exploit human detector models such as Feature Pyramid Networks~\cite{Lin2017}, and Faster R-CNN~\cite{Ren2015}. Papandreou et al.~\cite{Papandreou2017} propose one of the first top-down models which employ the Faster R-CNN for the person detector step and present a new representation for keypoints, which is a mixture of binary activation heatmap and the corresponding offset. The most recent top-down approaches which obtain SOTA accuracy are the Cascaded Pyramid Network (CPN) introduced by Chen et al.~\cite{Chen2018}, where multi-scale feature maps from different layers of the GlobalNet are integrated with an online hard keypoint mining loss for difficult-to-detect joints, and the model presented by Sun et al.~\cite{sun2020} that improves the heatmap estimation using high-resolution representations and multiple branches of different resolutions.

\subsubsection{Bottom-up:}
The recent architectures~\cite{Cao2021, Newell2017, Nie2018, Papandreou2018, li2020} take advantage of the confidence maps to detect the keypoints. Kreiss et al.~\cite{Kreiss2019} introduce a combination of confidence maps and vectorial parts for keypoints detection. Moreover, there are different approaches for encoding the part association used in the SOTA bottom-up models. OpenPose~\cite{Cao2021} introduces the Part Affinity Fields (PAFs) method to learn the body parts associations by encoding the location and direction, offset regression that uses the displacements of the keypoints~\cite{Nie2018, Papandreou2018}, and tag heatmap, which produces a heatmap as a tag for each keypoint heatmap~\cite{li2020, Newell2017}. Pose Partition Networks~\cite{Nie2018} present a dense regression approach over all the keypoints to generate individuals' partitions using the embedding spaces.

\begin{figure*}
\centering
\resizebox{.99\linewidth}{!}{\input{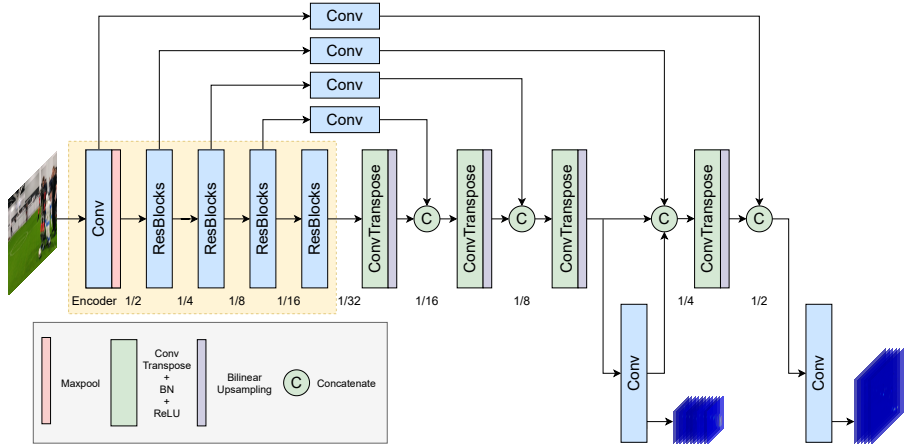}}
\caption{The architecture of the proposed single-stage encoder-decoder model. The model predicts heatmaps of both keypoints and limbs for scale $1/4$ and only heatmaps of keypoints for scale $1/2$, where each scale is supervised with an intermediate loss.}
\label{fig:2}
\end{figure*}

\section{Pose Estimation Model} \label{sec:methodology}

In this section, we present our real-time bottom-up approach to pose estimation of multiple humanoid robots. The aim is to predict the part coordinates and the part associations to build robot poses. In the following, we first describe the model, then explain the keypoint detection and the part association methods in detail.

\subsection{Network Architecture}
Following the successful results of NimbRo-Net~\cite{farazi2018nimbro} and NimbRo-Net2~\cite{rodriguez2019robocup}, we decided to utilize a similar architecture. This decision ensures that later we can combine this model with NimbRo-Net2 to have a unified network for multiple tasks related to the humanoid league. The proposed network is depicted in Fig.~\ref{fig:2}.

Our model is an encoder-decoder network which takes an RGB image of size $w \times h$. We observe that it is required to use a deeper encoder than the decoder to create a powerful feature extractor. The encoder is a pre-trained ResNet model~\cite{He2016}, in which the last fully connected and global average pooling layers are eliminated. The first layer is a $7\times 7$ convolutional with stride $2$, followed by a max-pooling layer. The rest of the encoder network consists of four modules of residual blocks with higher depths and lower resolutions as the number of modules increases. Each residual block consists of two or three convolutional layers, depending on the selected ResNet architecture, followed by batch normalization and ReLU activation and a shortcut connection. More fine-grained spatial information is present in the early layers, while in the final layers, the network extracts more semantic information.

In the decoder part, we utilize lateral connections from different parts of the encoder, which allows us to maintain the high-resolution information. For every lateral connection, we apply $1 \times 1$ convolution to generate a fixed number of channels. 
The decoder network has a feature pyramid structure involving four modules. At each level of the pyramid, the previous level's output is fed to the $3 \times 3$ transposed convolution followed by a bilinear upsampling to obtain a fixed number of higher-resolution features. These upsampled features are concatenated with the features from the corresponding lateral connection. Similar to the encoder, ReLU and batch normalization is used to get the final output of the module.

As high-resolution feature maps are essential for precise keypoint localization~\cite{Xiao2018}, we leverage two scales of the feature pyramid hierarchy, i.e., $1/4$ and $1/2$ resolutions. As a result, the keypoints heatmaps $\mathbf{\hat{K}^{s}}$ at each scale $s$ is generated by performing a final convolution on extracted features. As depicted in Fig.~\ref{fig:2}, we have two scales of keypoint heatmaps with intermediate supervision, inspired by HigherHRNet~\cite{Cheng2020}. The final keypoints heatmaps $\mathbf{\hat{K}}$ are the average over the predictions generated by these two scales after upsampling to the same resolution as the input image in order to achieve accurate high-resolution predictions. Note that only one scale of limbs heatmaps $\mathbf{\hat{L}}$ is utilized, as we observe that following the same approach as keypoints yields performance drop.

\subsection{Keypoint Detection and Part Association}

The ground truth heatmaps of the keypoints in a given image can be represented as the set $\mathbf{K} = \{\mathbf{K}_1, \mathbf{K}_2, ..., \mathbf{K}_P\}$, where  $\mathbf{K}_{p} \in \mathbb{R} ^ {w' \times h'}$, $p \in \{1, 2, ..., P\}$, and $P$ is the total keypoints of a robot instance, which is equal to six for our dataset (see Fig.~\ref{fig:3} (left)).
The heatmap $\mathbf{K}_{p}$ with the resolution of $w' \times h'$ includes the Gaussian heatmaps of the $p$th part of all the robot instances. Let $x_{p, n}$ and $y_{p, n}$ be the location of the $p$th part of the $n$th instance, where $n \in \{1, 2, ..., N\}$ and $N$ is the total number of existing robots with the visible $p$th part in an image. To embed the position of the annotated $p$th part, we use the 2D unnormalized Gaussian distribution with the center of $(x_{p, n}, y_{p, n})$ and the standard deviation $\sigma$, which is fixed for all the parts:

\begin{equation} \label{eqn:1}
    \mathbf{K}_{p}(x, y)=\exp(-\frac{(x-x_{p, n})^2 + (y-y_{p, n})^2}{2\sigma^{2}}).
\end{equation}
Due to occlusion or proximity of the robot instances in a given image, we utilize the pixel-wise max operation on Eq.~\ref{eqn:1} to preserve the Gaussian peak of the $p$th part for each instance.

For limbs, the ground truth heatmaps in a given image can be expressed as the set $\mathbf{L} = \{\mathbf{L}_1, \mathbf{L}_2, ..., \mathbf{L}_C\}$, where  $\mathbf{L}_{c} \in \mathbb{R} ^ {w' \times h'}$, $c \in \{1, 2, ..., C\}$, and $C$ is the total limbs of a robot instance that is five in this work (see Fig.~\ref{fig:3} (left)). Note that the intended utility of limbs is only to encode the relations between keypoints, so they do not necessarily lie on actual robot limbs. Therefore, to encode a limb's position, first, we compute a line segment between two keypoints and mark all of the points that lie on such limb, following the approach proposed by Li et al.~\cite{li2020}. Then having these offsets, the final Gaussian heatmap of each limb is generated by an unnormalized Gaussian distribution with the standard deviation $4\sigma$ that controls the spread of the Gaussian peak in the same way as for the keypoint heatmap.
The final limb heatmap is the average of all the robots' limb appearing in the image. In contrast to the PAFs method that encodes each limb in two channels as vector directions, we encode each limb type in a single channel. This simpler approach for encoding the limbs is enough for our application since our experiments show better performance than the PAF method.

\subsection{Loss}
We use the mean square error to compute the loss between the predicted heatmaps and the ground truth heatmaps for both keypoints and the limbs.
\begin{equation}
\label{eqn:2}
\mathcal{L}_{keypoints} = \frac{1}{2P}\sum_{s \in \{{\frac{1}{4}, \frac{1}{2}}\}}^{} \sum_{p=1}^{P} \mathbf{W} \cdot\left\|\mathbf{K}_{p}^{s} -\mathbf{\hat{K}}_{p}^{s}\right\|_{2}^{2},
\end{equation}
\begin{equation}
\label{eqn:3}
\mathcal{L}_{limbs} = \frac{1}{C}\sum_{c=1}^{C} \mathbf{W} \cdot\left\|\mathbf{L}_{c}-\mathbf{\hat{L}}_{c}\right\|_{2}^{2},
\end{equation}
where $W$ is a binary mask with $W=0$ when the annotation is missing in the image, and $s$ is the scale of the predicted heatmaps.
Finally, the total loss used to train the network is the sum of the keypoint loss (\ref{eqn:2}) and the limb loss (\ref{eqn:3}).

\subsection{Post Processing}

By performing Non-Maximum Suppression ($3 \times 3$ kernel) on the predicted keypoint heatmaps, we obtain the peak of each Gaussian heatmap and the location of its corresponding keypoint for the robot instances. We use the detected limb's heatmaps to acquire the candidate connections between the keypoints as~\cite{Cao2021}. As there are multiple robot instances in an image, it is required to group the keypoints to determine the poses corresponding to the correct individuals. Having the set of keypoints and the connection candidates, we employed the proposed greedy algorithm by Cao et al.~\cite{Cao2021} to solve the assignment problem and obtain the final pose of all robot instances. In this algorithm, instead of considering the fully connected graph, the goal is to obtain the minimum spanning tree of the pose instance and assign the adjacent tree nodes independently, resulting in a well-approximate solution with efficient computational cost.

\section{Dataset} \label{sec:dataset}
This section explains the paper's additional contribution, the HumanoidRobotPose dataset, including data collection and annotation procedures and the evaluation metrics used for this dataset.

Our goal was to collect a dataset containing both single and multiple robots to simulate the RoboCup's real conditions. We gathered many YouTube videos from the RoboCup Humanoid League, as well as some in-house videos and ROS bags. Some videos originate from the qualification videos, which only demonstrate a specific robot; therefore, they only consist of a single pose. To include videos with multiple robots and increase the diversity of robots in the dataset, we also employ videos from drop-in games and round-robin competitions. These videos are from recent years and contain various view angles, lens distortions, brightness, and robots. Note that in most of the videos, there are humans present in the pictures, e.g., the robot handler, the referee, and audiences around the field.
Overall, we annotated over $1.5K$ manually selected frames from $23$ videos with around $2.3$k robot instances. These frames include teen and adult sized robots and contain more than ten different robot types. About $30$ percent of the dataset was exclusively used for testing. Note that testing frames were collected from different videos than the training videos.

\begin{figure}[t]
  \centering
  \begin{minipage}{0.47\textwidth}
    \centering
    \vspace{-6px}
    \includegraphics[width=1.\linewidth]{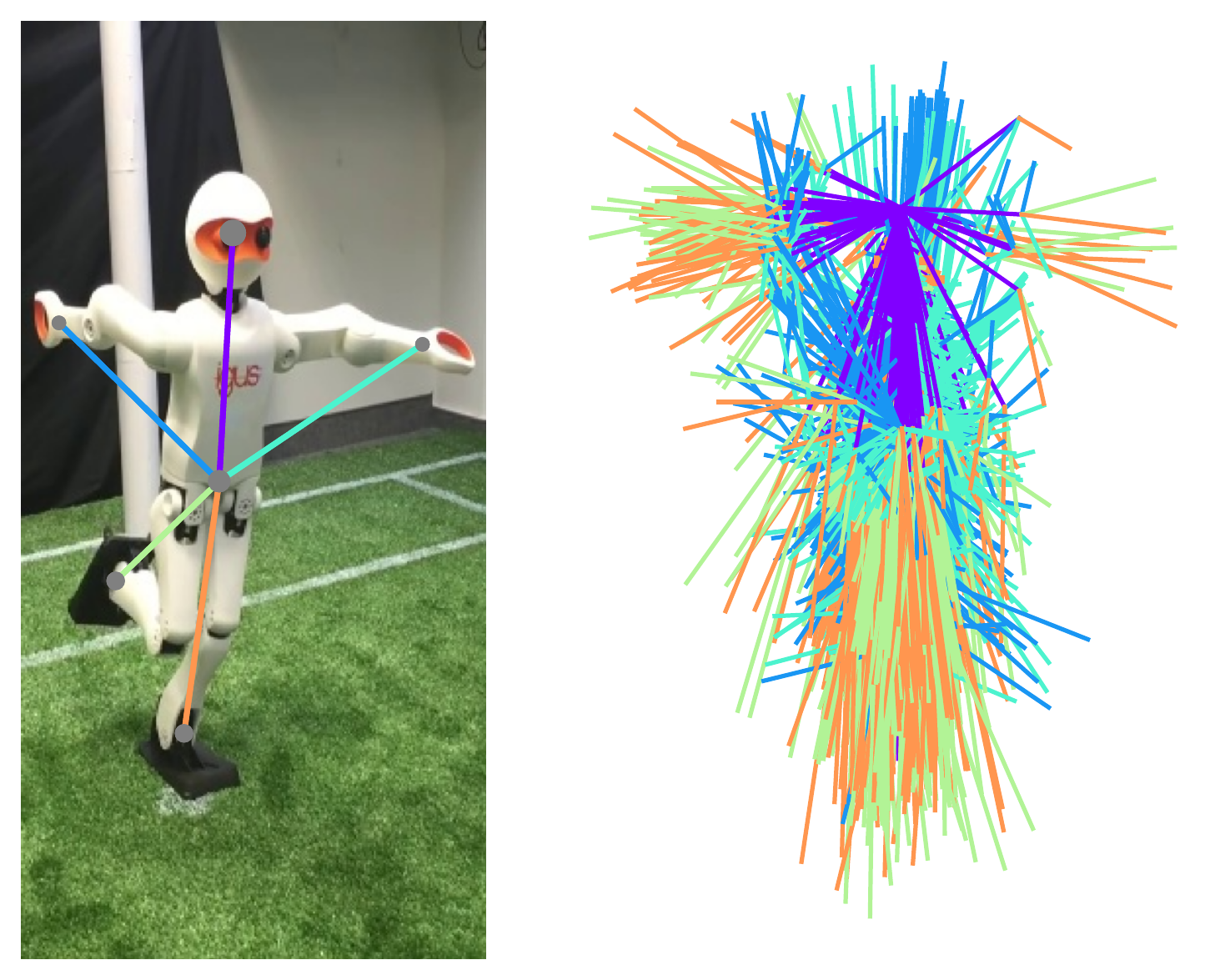}
    \captionof{figure}{An example of the annotations depicted in left. Visualization for the distribution of pose variability by considering the head as the origin is shown in right. Note that we did not choose trunk joint as the origin to show the dataset's variability better.}
    \label{fig:3}
  \end{minipage}
  \hspace{0.04\textwidth}
  \begin{minipage}{0.47\textwidth}
    \centering
    \includegraphics[width=1.\linewidth, height=1.8in]{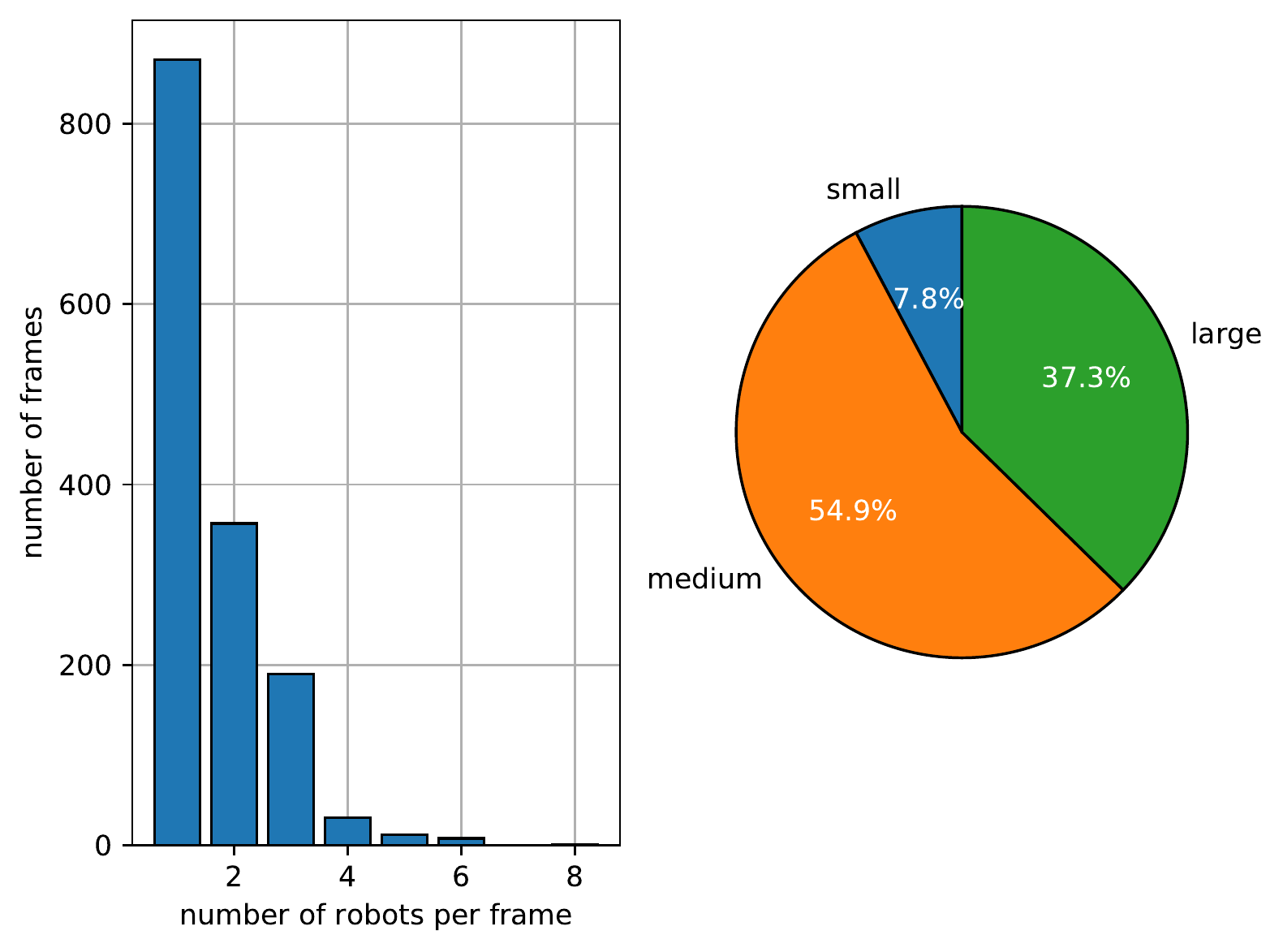}
    \captionof{figure}{Statistics for our dataset. The diversity of the number of robot instances is illustrated in left, and the scale proportions of robot instances is shown in right. The definition of small, medium and large scale is identical to the COCO dataset.}
    \label{fig:4}
  \end{minipage}
\end{figure}

\subsection{Data Annotation}
For data annotation, we used Supervise.ly\footnote{https://supervise.ly}, a web-based data annotations and management tool. We decided to ignore the truncated or severely occluded points in the image, which are usually considered invisible keypoints. For each robot, six keypoints are annotated, including head, trunk, hands, and feet. The head keypoint is important, for instance, to estimate the height of the goalie robot. We use these few keypoints to avoid annotation costs; however, they can be easily extended to more keypoints. We define a minimal pose representation by five limbs from these keypoints, which would be sufficient for the current soccer behavior level. The annotation for a robot instance is illustrated in Fig.~\ref{fig:3} (left).

To show the diversity of our dataset, we visualize the variability of annotated poses in Fig.~\ref{fig:3} (right) and statistics of the number and scales of the robot instances are presented in Fig~\ref{fig:4}. About $60$ percent of the collected frames contain single pose instances and robot instances with medium scale, i.e., $[32^2 <$ segment area $< 96^2]$, where the segment area of a robot instance is measured using the size of the minimum encapsulating rectangle of the annotated keypoints. Our definition of the size scales is identical to the COCO dataset~\footnote{\label{cocoeval}https://cocodataset.org/\#keypoints-eval}.

\subsection{Evaluation Metrics}
We use the Object Keypoint Similarity (OKS) metric from COCO keypoint dataset~\cite{Ronchi2017}. The OKS of a robot $r$ between the detected keypoint ($\hat{y}_{i, r}$) and its corresponding ground truth ($y_{i, r}$) can be written as follows:

\begin{equation}
\label{eqn:4}
    OKS(\hat{y_r}, y_r)=\frac{\sum_{i}
    e^{-\frac{\|\hat{y}_{r_i}-y_{r_i}\|_{2}^{2}}{2 a^{2} k_{i}^{2}}}\delta(v_{i}>0)}{\sum_{i}\delta(v_{i}>0)},
\end{equation}
where $k_i$ is a constant specific to each keypoint, which is equal for all keypoints in our dataset, $a$ is the segment area of the robot instance measured in pixels, and $v_i$ is the keypoint visibility flag in the ground truth ($v_i = 0$ for the invisible keypoint). The OKS metric is robust to the number of visible keypoints as it gives equal importance to the robot instances with different numbers of visible keypoints.
The evaluation metrics used for the proposed dataset are as following: AP (the mean average precision over 10 OKS thresholds = [0.50:0.05:0.95]), AP\textsuperscript{50} (AP at OKS threshold = 0.50), AP\textsuperscript{75}, AP\textsuperscript{M} for medium scale robot instances, AP\textsuperscript{L} for large scale instances, and AR (the mean of average recall over 10 OKS thresholds).

\section{Experiments}

\begin{figure}[!ht]
\centering
\includegraphics[width=\linewidth]{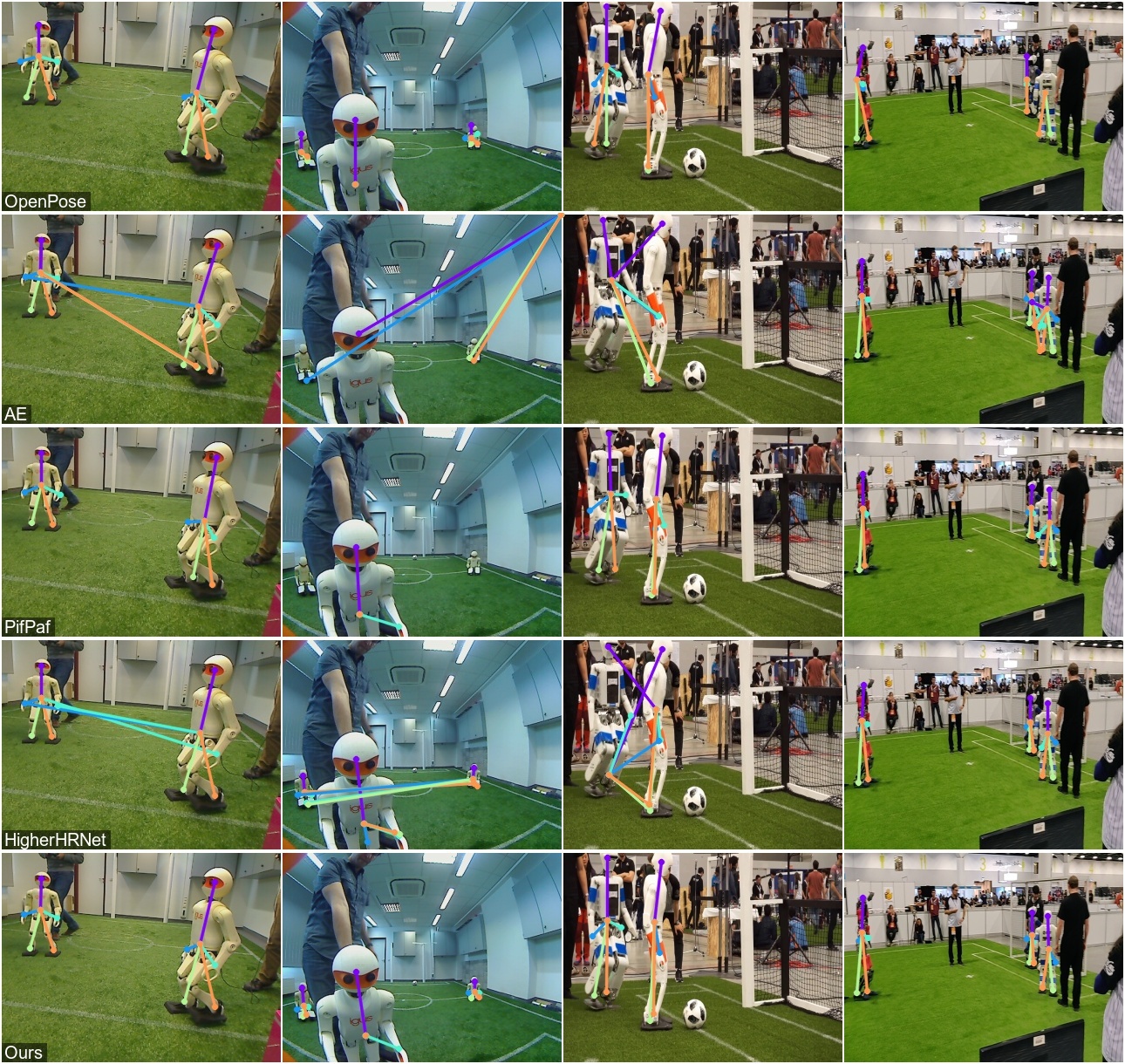}
\caption{Four sample results from the test set. Methods used from top to bottom: OpenPose~\cite{Cao2021}, AE~\cite{Newell2017}, PifPaf~\cite{Kreiss2019}, HigherHRNet~\cite{Cheng2020}, and our model.}
\label{fig:5}
\end{figure}

We compare the proposed method with SOTA bottom-up approaches on the HumanoidRobotPose dataset. These approaches are OpenPose~\cite{Cao2021}, Associative Embedding (AE)~\cite{Newell2017}, PifPaf~\cite{Kreiss2019}, and HigherHRNet~\cite{Cheng2020}. OpenPose~\cite{Cao2021} utilizes confidence maps to localize the keypoints and PAFs to encode the body parts' location and orientation. For grouping the detected keypoints, the greedy algorithm is proposed in which each part is scored, computing the line integral on the corresponding PAF. Associative Embedding (AE)~\cite{Newell2017}, merges the stacked hourglass architecture~\cite{Newell2016} with associative embedding. 

PifPaf~\cite{Kreiss2019} proposes Part Intensity Field to detect and localize the keypoints and Part Association Fields to associate body parts with each other.

HigherHRNet~\cite{Cheng2020} is using an adopted top-down model as the backbone with a transposed convolution module to predict higher resolution heatmaps for the keypoints detection. Similar to the AE approach, in HigherHRNet, the associative embedding is employed to parse the poses. Following the configurations provided by the original papers, we reported the details of models and the inference time evaluated on NimbRo-OP2X robot hardware~\cite{ficht2018nimbro} in Table~\ref{tab1}.

For all methods, the hyperparameters are tuned to achieve the best possible results. Our model is trained using the AdamW~\cite{loshchilov2019} optimizer with learning rate of $10^{-4}$, batch size 16 and weight decay of $10^{-4}$ for the total $200$ epochs. Note that the encoder is initialized by pre-trained ResNet weights on ImageNet. We conduct data augmentation that includes random horizontal flip, random rotation, random scaling, and random translation during training. 
% Then, the augmented image is resized to $384 \times 384$. Note that during the inference of our model, the input size can also be non-square.

\begin{table*}[t]
\centering
\caption{Details for the used methods.}
\label{tab1}
\begin{tabular}{c|c|c|c|c|c}
\hline
Method & Input Size & Backbone & Params & GFLOPs & FPS \\
\hline
OpenPose~\cite{Cao2021} & 368 & VGG19 & 25.8M & 159.8 & 14 \\
AE~\cite{Newell2017} & 512 & Hourglass & 138.8M & 441.6 & 5 \\
PifPaf~\cite{Kreiss2019} & 385 & ShuffleNetV2 & \textbf{9.4M} & 46.3 & 13 \\
HigherHRNet~\cite{Cheng2020} & 512 & HRNet-W32 & 28.6M & 94.7 & 13 \\
\hline
Ours & 384 & ResNet18 & 12.8M & \textbf{28.0} & \textbf{48} \\
\end{tabular}
\end{table*}

\begin{table*}[t]
\centering
\caption{Results on the test set.}
\label{tab2}
\begin{tabular}{c|c|c|c|c|c|c|c|c|c|c}
\hline
Method & AP & AP\textsuperscript{50} & AP\textsuperscript{75} & AP$^{M}$ & AP$^{L}$ & AR & AR\textsuperscript{50} & AR\textsuperscript{75} & AR$^{M}$ & AR$^{L}$ \\
\hline
OpenPose~\cite{Cao2021} & 67.9 & 80.0 & 70.0 & 73.8 & 73.1 & 68.7 & 80.1 & 70.4 & 74.8 & 74.4 \\
AE~\cite{Newell2017} & 62.9 & 71.9 & 64.1 & 64.0 & 72.9 & 64.6 & 73.9 & 65.6 & 64.7 & 76.0 \\
PifPaf~\cite{Kreiss2019} & 76.1 & 81.6 & 75.6 & 76.0 & \textbf{91.0} & 77.9 & 83.6 & 77.2 & 77.7 & \textbf{93.0} \\
HigherHRNet~\cite{Cheng2020} & 73.4 & 84.1 & 75.6 & 80.3 & 78.7 & 76.2 & 85.3 & 77.2 & 81.4 & 83.0 \\
\hline
Ours & \textbf{78.1} & \textbf{84.6} & \textbf{79.6} & \textbf{87.5} & 80.2 & \textbf{79.4} & \textbf{85.4} & \textbf{80.6} & \textbf{88.4} & 81.6 
\end{tabular}
\end{table*}

The results on the test set are reported in Table~\ref{tab2}. The reported results are achieved without performing the flip test or the multi-scales test for preserving the methods to be real-time. Our proposed method with ResNet18 backbone outperforms the best existing methods in all metrics except for large scale when we train the models from scratch on our dataset (see Table~\ref{tab2}). Note that compared to other baselines, our model can utilize our limited dataset better. Based on AP results of medium and large scales, our model can better handle the different scales than the other approaches. Moreover, the strict metric results demonstrate that the predicted pose instances are more accurate compared to the other methods due to the high-resolution predictions. Fig.~\ref{fig:5} illustrates some samples of estimated poses for all the approaches.

\section{Ablation Study}

This section investigates different backbones for the encoder part of our model and the importance of employing multi-scale predictions in our approach. As shown in Table~\ref{tab3}, although applying a deeper encoder helps achieve better performance, it negatively affects the inference time of the model. Moreover, AP results demonstrate that without multi-scale heatmaps, the accuracy of predicted keypoints drops.

\begin{table*}
\centering
\caption{Ablation study: the effectiveness of backbones and multi-scale predictions on the test set.}
\label{tab3}
\begin{tabular}{c|c|c|c|c|c|c|c}
\hline
Backbone & Multi-scale & AP & AP$^{M}$ & AP$^{L}$  & Params & GFLOPs & FPS\\
\hline
ResNet18 &  & 76.6 & 83.6 & 82.0 & \textbf{12.4M} & \textbf{17.8} & \textbf{83} \\
ResNet18 & \checkmark & 78.1 & \textbf{87.5} & 80.2 &  12.8M & 28.0 & 48 \\
ResNet50 & \checkmark & \textbf{78.6} & 86.0 & \textbf{82.1} &  27.0M & 47.7 & 25 \\
\end{tabular}
\end{table*}

\section{Conclusion}
In this paper, we presented a lightweight bottom-up model for estimating multiple humanoid robot poses in real-time. We showed that our proposed model is capable of multi-robot pose estimation on NimbRo-OP2X robot hardware and is more suitable for the RoboCup humanoid league in comparison with other SOTA models. For the future, we will use this model for advanced soccer behavior decisions like recognizing rival robots' actions or anticipating the ball's movement direction before the kicking motion. Since the developed model is very similar to NimbRo-Net2, we will combine them to produce a unified network for diverse perception tasks in RoboCup.

{\footnotesize \paragraph{Acknowledgment:}
\label{acknowledgment}
This work was partially funded by grant BE 2556/16-2 (Research Unit FOR 2535 Anticipating Human Behavior) of the German Research Foundation (DFG).
}

\bibliographystyle{splncs04}
\bibliography{references}

\end{document}